%% file: template.tex
\documentclass{article}

\usepackage{arxiv}

\usepackage[utf8]{inputenc} 
\usepackage[T1]{fontenc}    
\usepackage{hyperref}       
\usepackage{url}            
\usepackage{booktabs}       
\usepackage{amsfonts}       
\usepackage{nicefrac}       
\usepackage{microtype}      
\usepackage{lipsum}
\usepackage{amsmath}
\usepackage{graphicx}
\usepackage{orcidlink}
\graphicspath{ {./images/} }

\title{A Human-in-the-Loop Deep Learning Framework for Color Reconstruction of Lenticular Films}

\author{
Saptarshi Neil Sinha\,\orcidlink{0000-0001-6637-0379} \\
Virtual and Augmented Reality, Fraunhofer IGD, Darmstadt, Germany \\
\texttt{saptarshineilsinha@gmail.com}
\And
Tiago Kleist\,\orcidlink{0009-0005-0961-7036} \\
Virtual and Augmented Reality, Fraunhofer IGD, Darmstadt, Germany
\\
\texttt{tiago.kleist@igd.fraunhofer.de}
\And
Giorgio Trumpy\,\orcidlink{0000-0001-9534-0507} \\
Colourlab, Norwegian University of Science and Technology, Gjøvik, Norway \\
\texttt{giorgio.trumpy@ntnu.no}
}

\begin{document}
\input{text/our_macros}
\maketitle
\input{text/abstract}
\input{text/introduction}
\input{text/related_works}
\input{text/methodology}
\input{text/evaluation}
\input{text/conclusion}

\input{text/acknowlegements}

\bibliographystyle{unsrt}  
\bibliography{dolce}  

\end{document}

%% file: text/our_macros.tex
\def\ie{i.e.\ }
\def\eg{e.g.\ }
\def\etal{et~al.\ }
\def\wrt{w.r.t.\ }


\newcommand{\todo}[1]{\textcolor{red}{#1}}
\newcommand{\nrevision}[2]{{\color{red}\sout{#1}}{\color{blue}\uwave{#2}}}
\newcommand{\grevision}[2]{{\color{red}\sout{#1}}{\color{orange}\uwave{#2}}}

%% file: text/abstract.tex
\begin{abstract}
Historical lenticular films, such as those created with the Kodacolor process, encode color information in a distinctive spatial format. This structure requires specialized techniques for accurate color reconstruction. While recent signal processing approaches like doLCE and deep learning methods like deep-doLCE have advanced automated color recovery, they often fail with cases such as curved lenticules, low-contrast, or badly captured regions. We propose a human-in-the-loop (HITL) deep learning framework which is designed for color reconstruction in lenticular films. Our approach introduces an editable, vector-based representation of lenticule boundaries, allowing experts to interactively refine boundary positions before color extraction and demosaicing. This decoupled architecture enables targeted corrections and iterative fine-tuning, embedding expert knowledge into the detection model and improving robustness across challenging frames. To preserve image details using information solely present in the original silver emulsion, we merge the reconstructed chrominance with the original film scan’s luminance. We evaluate our pipeline on a challenging lenticular film sequence where previous automated approaches fail and the reconstructed colors are not suitable for exhibition. In contrast, our HITL approach successfully produces high-quality, exhibitable color reconstructions with preserved texture. This work is the first to combine expert guidance, editable intermediate representations, and texture-preserving post-processing for lenticular film color reconstruction, advancing the state of the art in this field.
\keywords{Color reconstruction \and Lenticular films \and Historical photographs \and Deep Learning \and Segmentation \and Human-in-the-Loop training}
\end{abstract}

%% file: text/introduction.tex
\section{Introduction}
\label{sec:introduction}
\begin{figure*}[htb!]
    \centering
    \includegraphics[width=0.8\linewidth]{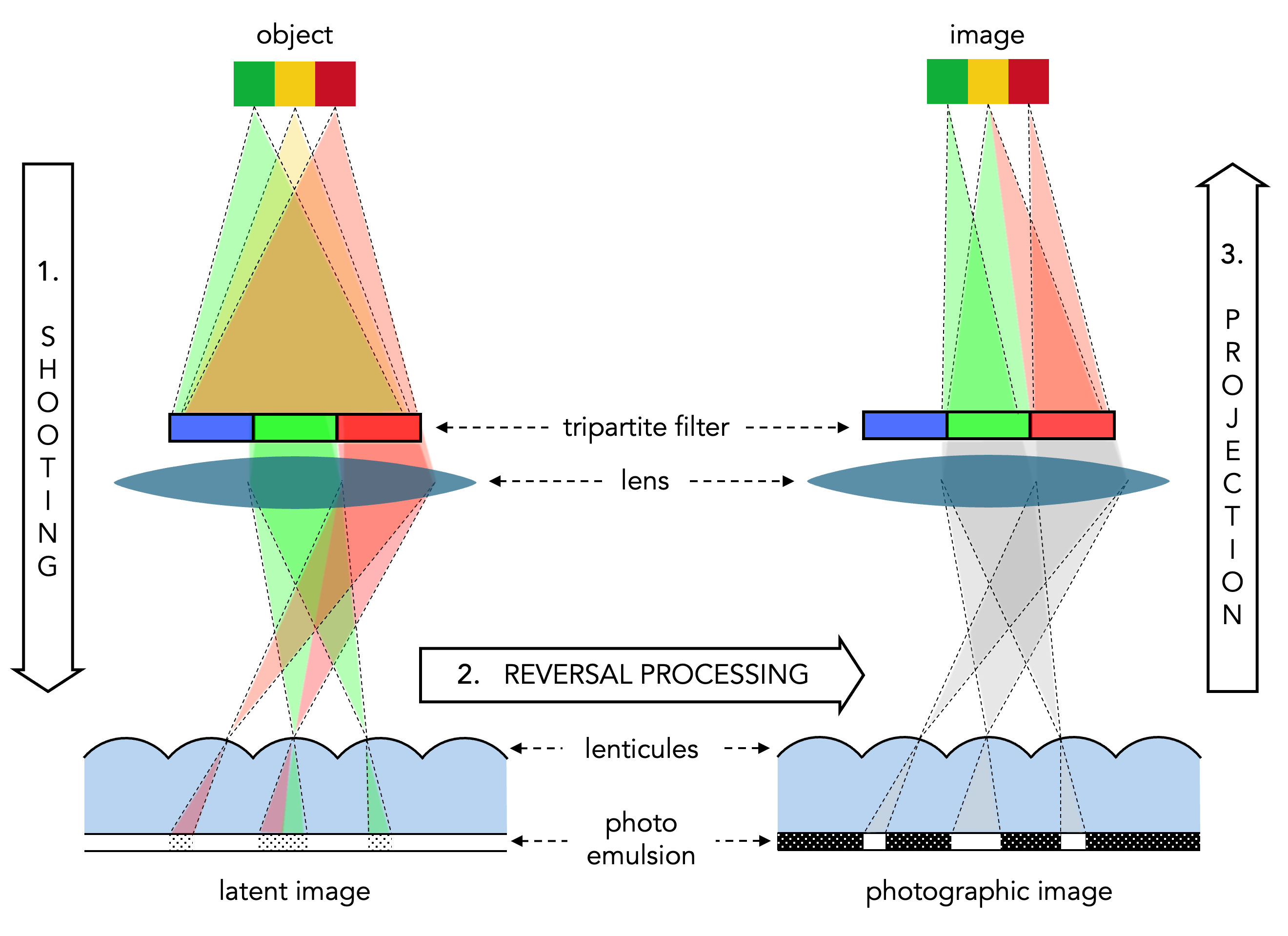}
    \caption{Diagram describing the three stages of the lenticular color reproduction: shooting/processing/projection. The ‘object’ is represented by three color elements green, yellow and red.}
     \label{fig:lentproc}
\end{figure*}

Historical films provide invaluable insights into the past and serve as essential records for cultural heritage, research, and preservation. Before the advent of modern color film, black-and-white motion pictures dominated the market worldwide. The demand for realistic color led to numerous experimental processes, but few were commercially successful. In the late 1920s, Eastman Kodak introduced the 16mm Kodacolor process, the first amateur color motion picture system based on lenticular film technology. This innovation allowed home filmmakers to capture color movies at a time when alternatives were limited and costly. Unlike conventional film, lenticular film use a series of fine, parallel cylindrical lenses called lenticules embossed on the film’s surface. Behind each lenticule, a black-and-white emulsion records the image, and a color filter in front of the camera lens splits the light into red, green, and blue components. These are encoded as grayscale stripes under each lenticule. Viewing the color image requires a special projector with a matching filter and lens system, aligning the viewing geometry precisely with the original camera setup. A diagram of the Kodacolor process is reported in Fig. \ref{fig:lentproc}.

For several years, lenticular processes like Kodacolor dominated the amateur color film market. However, in 1935, Kodak introduced Kodachrome, a revolutionary film with integral multilayer emulsions that allowed direct color recording and standard projection. Kodachrome quickly replaced lenticular film in the market and maintained its dominance for many years. During this period, magnetic tape emerged as a parallel technology, enabling analog video recording through formats such as VHS and Betamax and reshaping the consumer imaging landscape. In the 1970s, Bayer filters emerged and eventually became the leading technology for color imaging, surpassing Kodachrome. Today, as analog projection devices become obsolete, it is increasingly important to digitally preserve lenticular films and reconstruct their colors faithfully.

Recent developments in lenticular film color reconstruction have centered around automated pipelines. The signal-processing-based doLCE framework \cite{ReutelerGschwind2014RundbriefFotografie} and its deep learning extension, deep-doLCE \cite{daronco2022deeplearningapproachdigital}, have advanced the field by segmenting lenticule boundaries and interpolating missing color information. However, these approaches are limited by their reliance on fully automatic detection, which can fail in challenging cases, such as curved lenticules, low-contrast regions, physical damage, or poor capture conditions, often leading to reconstruction artifacts or frames rendered unsuitable for public exhibition.

To overcome these limitations, we propose a novel human-in-the-loop (HITL) deep learning pipeline for lenticular color reconstruction. Our approach introduces an editable intermediate representation of lenticule boundaries, enabling experts to interactively refine boundary positions and correct errors before color extraction and demosaicing. This decoupled architecture allows targeted corrections, ensuring higher fidelity in difficult cases. Expert corrections are used to iteratively fine-tune the detection model, embedding domain knowledge and improving robustness across the sequence. Furthermore, a post-processing step merges the reconstructed chrominance with the original scan's brightness, preserving the authentic level of detail of the silver emulsion. In light of these changes, our main contributions are as follows:
\begin{itemize}
    \item We introduce the first editable representation of lenticule boundaries in the deep learning based lenticular color reconstruction pipeline, allowing flexible expert-driven corrections that are not possible with previous approaches such as deep-doLCE~\cite{daronco2022deeplearningapproachdigital}.
    \item We present a human-in-the-loop (HITL) architecture for lenticular color reconstruction, enabling experts to review, correct, and iteratively improve the detection and colorization process. 
    \item We propose an iterative fine-tuning strategy in which expert-corrected lenticule boundary annotations are used to progressively adapt the detection network to film-specific 
    color reconstruction failures, embedding domain knowledge directly into the model and improving reconstruction quality across the entire sequence.
\end{itemize}
To the best of our knowledge, this is the first work to introduce a human-in-the-loop approach for lenticular color reconstruction in historical films, combining editable intermediate representations, expert-driven correction, and texture-preserving post-processing.

%% file: text/related_works.tex
\section{Related works}
\label{sec:related_works}
Image colorization is an underconstrained problem, as a single grayscale
image can have many plausible colorizations~\cite{gray_scale_colorization_4}.
Early approaches addressed this through user-guided scribble-based
methods~\cite{levin2004colorization}, reference-based color
transfer~\cite{welsh2002transferringcolor, irony2005colorization}, and
optimization-based propagation. Deep learning methods later automated color
prediction using end-to-end CNNs~\cite{cheng2015deepcolorization,
iizuka2016lettherecolor, larsson2016learning}, user-guided
networks~\cite{zhang2017realtimecolorization}, generative
models~\cite{vitoria2020chromagan, gray_scale_colorization_3}, and
exemplar-based approaches~\cite{su2020instanceawarecolorization}. Modern
tools such as DeOldify~\cite{deoldify} achieve strong automatic results but
remain prone to hallucinated or implausible
colors~\cite{gray_scale_colorization_1}.
Unlike natural image colorization, lenticular film reconstruction is not
free to assign arbitrary colors, as the color information is spatially encoded across
horizontal sub-stripes inside each lenticule. This makes naive colorization risky and unreliable unless this encoding is respected. The pioneering doLCE work~\cite{ReutelerGschwind2014RundbriefFotografie} introduced a practical two-stage pipeline within a single processing framework. First, the method detects lenticule boundaries by analyzing the intensity profile and selecting the local minima, addressing the boundary-detection problem. Then, it reconstructs color by propagating a single RGB triplet across the width of each lenticule, similar to a constrained demosaicing step. This approach ensures that the colorization process respects the spatial encoding structure inherent to lenticular films. Building on this, D’Aronco et al. proposed a deep-learning extension, deep- doLCE~\cite{daronco2022deeplearningapproachdigital}, where a segmentation network localizes lenticule boundaries and a separate colorization network interpolates the missing color information inside each lenticule, enforcing a truthful colorization that adheres to the encoded horizontal color structure. However, deep-doLCE~\cite{daronco2022deeplearningapproachdigital} can still fail on challenging cases, such as curved lenticules, low-contrast regions, physical damage, or poor capture conditions. In these situations, automatic detection fails, often requiring ad-hoc manual intervention or leading to the affected frames being discarded.
The limitations of fully automatic approaches have driven the development of human-in-the-loop (HITL) paradigms in computer vision and creative AI. Works like Draw-with-Me~\cite{draw_with_me} demonstrate that interactive, human-guided restoration can outperform fully automatic methods by preserving texture and perceptual fidelity. Human-in-the-loop training has applications and benefits across different domains~\cite{survey_human_in_the_loop}, with concrete successes in domain-specific segmentation (e.g., underwater coral imagery with foundation-model priors and sparse labels \cite{coral_seg_human}), video annotation that combines active sample selection with test-time refinement \cite{video_seg_human}, and interactive atlas-based segmentation of 3D assets for cultural heritage and content production workflows~\cite{kuhn2026humanintheloopatlasbased3dasset}. Foundational models such as Segment Anything \cite{kirillov2023segany} and DINOv2 \cite{oquab2023dinov2} provide robust priors that empower HITL workflows, while papers on HITL for machine creativity~\cite{chung2021humanloopmachinecreativity} discuss curatorship and collaboration as pathways to richer, multi-modal outcomes. In the context of personalized, HITL-guided image generation, multi-round human feedback can steer diffusion models toward user targets without task-specific training~\cite{Rajagopalan2026PersonalizedImageGeneration}.
We take inspiration from human-in-the-loop (HITL) approaches in various domains, including image restoration, where expert intervention consistently improves results. Hence, we extend deep-doLCE by incorporating HITL strategies that facilitate more effective boundary editing and the integration of expert priors, leading to enhanced boundary detection and colorization.

%% file: text/methodology.tex
\section{Methodology}
\label{sec:methodology}
\begin{figure*}[htb!]
    \centering
    \includegraphics[width=\linewidth]{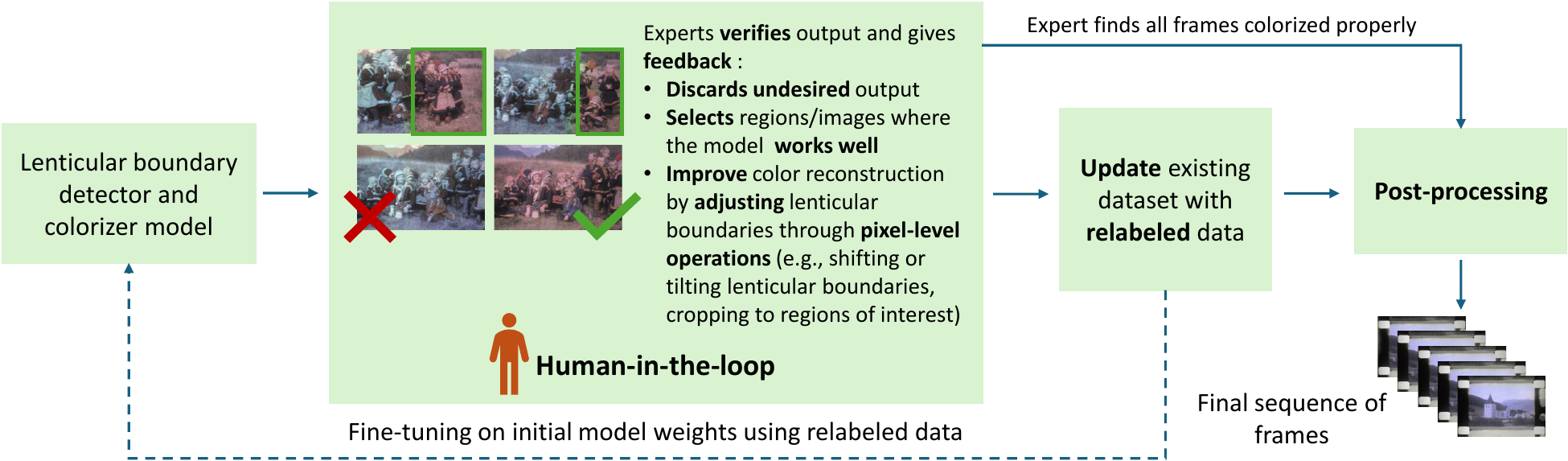}
    \caption{Overview of our human-in-the-loop pipeline. An initial lenticular boundary detector and colorization model~\cite{daronco2022deeplearningapproachdigital} processes the scanned film. Experts then review the results, discarding undesired outputs, selecting reliable regions, and improving color reconstruction by interactively adjusting lenticular boundaries (e.g., shifting, tilting, cropping to regions of interest). The corrected annotations are used to update the dataset and fine-tune the initial model. This process can be repeated, recursively embedding expert priors into the model. Once the expert confirms that all frames are properly colorized, post-processing is applied to produce the final sequence of frames.}
    \label{fig:pipeline_doLCE2.0}
\end{figure*}

In this section we describe our human-in-the-loop pipeline for color reconstruction of lenticular films. We first summarize the underlying two-stage deep learning architecture introduced by D'Aronco et al.~\cite{daronco2022deeplearningapproachdigital} and how we incorporated our work in this pipeline (Figure~\ref{fig:dolce1_with_user_guidance}). We then explain how we utilize this editable lenticular boundary representation in an iterative expert-in-the-loop refinement workflow (Figure~\ref{fig:pipeline_doLCE2.0}).
\subsection{Automated color reconstruction pipeline}
\label{subsec:automated_color_reconstruction}
\begin{figure*}[htb!]
    \centering
    \includegraphics[width=\linewidth]{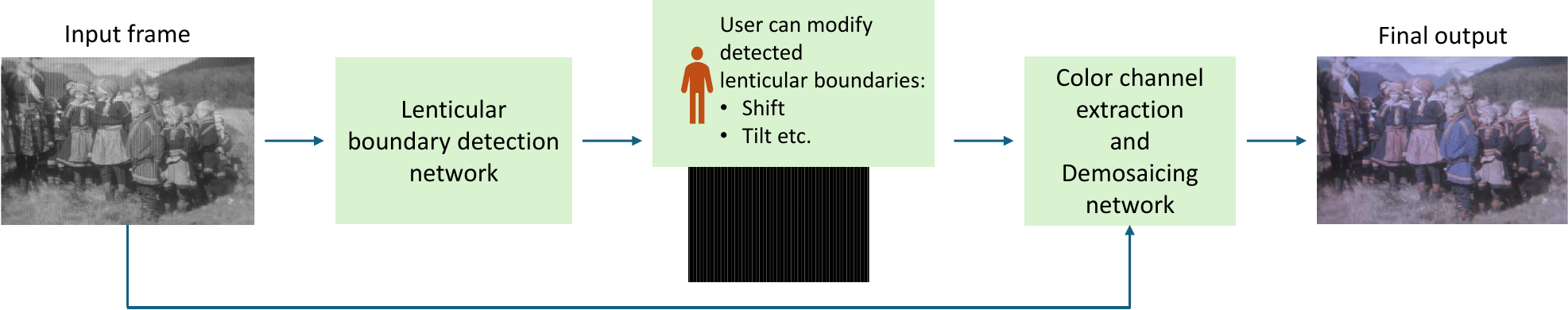}
    \caption{Original two-stage pipeline~\cite{daronco2022deeplearningapproachdigital} for color reconstruction, as used in our work with decoupled lenticular boundary detection and colorization. The input frame is first processed by a lenticular boundary detection network, whose output is exposed as an editable boundary layer that the user can modify (e.g., shift, tilt) before color reconstruction. The refined boundaries are then passed to the color channel extraction and demosaicing network to produce the final color output. This editable intermediate representation is new in our approach and enables human-in-the-loop correction (see Figure~\ref{fig:pipeline_doLCE2.0}) of the lenticules.}
    \label{fig:dolce1_with_user_guidance}
\end{figure*}

Our pipeline builds on the two-stage deep learning approach (see Figure~\ref{fig:dolce1_with_user_guidance}) introduced in \cite{daronco2022deeplearningapproachdigital} for automated color reconstruction of scanned lenticular films. The authors propose a two-stage, deep-learning-based pipeline to reconstruct color from scanned grayscale lenticular films.

In the first stage, they detect the lenticule boundaries. A U-Net~\cite{unet} (with ResNet~\cite{resnet} encoder) is trained as a binary segmentation model to label pixels as “boundary” vs “inside lenticule.” The network is trained using successful doLCE outputs~\cite{ReutelerGschwind2014RundbriefFotografie} (a signal-processing-based method that detects lenticules by analyzing the image intensity profile) as pseudo-ground truth. Small random rotations are applied so the network learns to handle slightly tilted lenticules. The resulting raster boundary map is then refined into a vector representation by fitting straight lines (one per lenticule) across the image, using an optimization that aligns these lines with high-probability boundary pixels while regularizing the average lenticule width and its smooth variation across the frame. From these fitted boundaries, they extract the grayscale samples corresponding to the red (R), green (G), and blue (B) sub-stripes inside each lenticule, assembling a narrower “striped” RGB image in which each column contains only one known color channel.

In the second stage, they colorize this striped image using another U-Net-like network trained on ordinary RGB images (DIV2K~\cite{Agustsson_2017_CVPR_Workshops}) that have been artificially converted to the same striped pattern by zeroing out two channels per column. To preserve historical color fidelity this network does not predict arbitrary colors. For each pixel and each channel it predicts non-negative interpolation weights that sum to one for a fixed set of nearby valid samples of the same color effectively learning how to interpolate rather than hallucinate colors. The training approach uses a combination of L1 and adversarial losses to produce sharp, realistic results, but it can still fail on difficult cases such as strongly tilted lenticules, large dark regions, or physical damage. In that workflow, such failures typically required ad-hoc manual intervention or were discarded.
\subsection{Decoupled architecture for interactive refinement}
\label{subsec:decoupled_architecture}
Our pipeline (see Figure~\ref{fig:dolce1_with_user_guidance}) explicitly decouples the lenticular boundary detection and colorization modules of the original architecture~\cite{daronco2022deeplearningapproachdigital}, and exposes the detected boundaries as an editable intermediate representation. This allows users to adjust the lenticular boundaries (e.g., shift (see Figure~\ref{fig:dolce2_shift}) or tilt(see Figure~\ref{fig:dolce2_tilt}) them) before the colorization stage, enabling a human-in-the-loop refinement process (see Figure~\ref{fig:pipeline_doLCE2.0}) that mitigates the failure cases described in Subsection~\ref{subsec:automated_color_reconstruction}. In the original pipeline, by contrast, detection and colorization were tightly coupled, and the colorization network operated directly on the (fixed) detected boundaries, leaving no possibility for intermediate correction.

In our new design, the lenticule boundary detection module outputs an \emph{editable} vector representation that can be iteratively adjusted and evaluated before colorization is approved (see Figure~\ref{fig:dolce1_with_user_guidance}). Each boundary is stored as a parameterized line defined by its top and bottom coordinates, which allows experts to manipulate the lenticule grid directly. The interface supports geometric edits such as rotating the grid to correct global tilt (see Figure~\ref{fig:dolce2_tilt}), horizontally shifting lenticular boundaries (see Figure~\ref{fig:dolce2_shift}) to fix positional offsets (e.g., in low-contrast regions), cropping to reliable regions when full color reconstruction is not possible  (see Figure~\ref{fig:user_refinement}), and discarding frames which cannot be corrected by these operations. Since, detection and colorization are decoupled, all these edits propagate cleanly to the subsequent color extraction and reconstruction without re-running or retraining the detection network. Once the expert approves the boundary positions, the corrected vectorial lenticule map is used for fine-tuning the deep learning model (see Figure~\ref{fig:pipeline_doLCE2.0}). 
\subsection{Human-in-the-loop refinement framework}
We introduce a human-in-the-loop refinement pipeline specifically designed for the color reconstruction of lenticular film. Rather than relying solely on automated processing, this framework integrates expert knowledge at key stages to iteratively improve reconstruction quality. The overall workflow is illustrated in Figure~\ref{fig:pipeline_doLCE2.0}, and each stage is described in detail in the following subsections.
\subsubsection{Initial batch processing}
All frames of a lenticular film are first processed through the automated baseline pipeline described in Subsection~\ref{subsec:automated_color_reconstruction} to produce initial colorized outputs. This provides a baseline reconstruction for the entire film sequence, which typically succeeds for the majority of frames.
\begin{figure*}[htb!]
    \centering
    \includegraphics[width=0.8\linewidth]{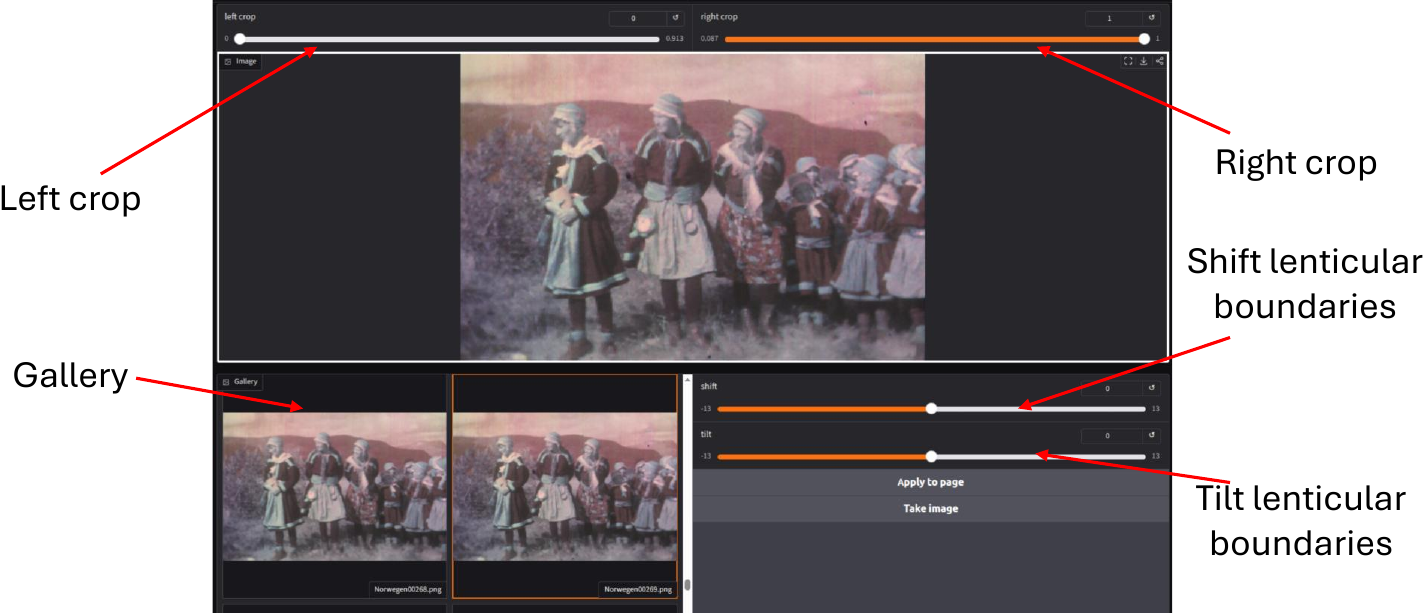}
    \caption{User interface for expert-guided lenticular boundary annotation and color reconstruction verification.}
     \label{fig:hitl_ui}
\end{figure*}
\begin{figure*}[htb!]
    \centering
    \includegraphics[width=\linewidth]{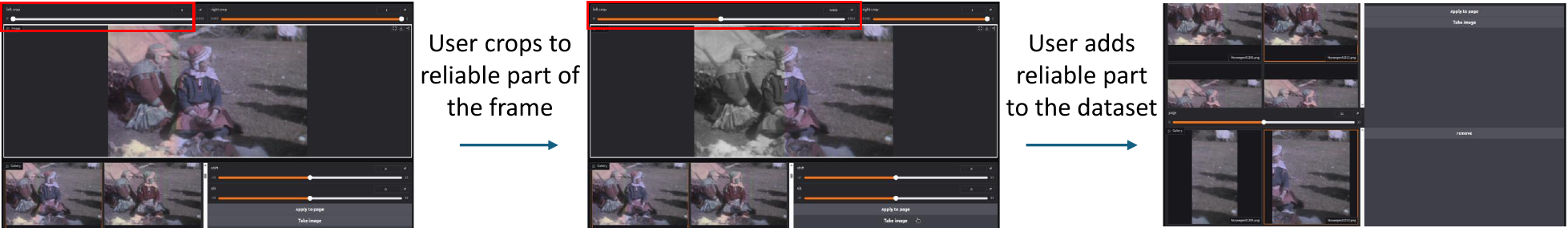}
    \caption{\textit{Crop/Select region of interest}: (Left) The original 
reconstructed frame with discolored regions from deep-doLCE output. (Middle) The user crops to the reliable part of the frame, with the uncropped region visualized in grayscale. (Right) The user adds the reliable cropped part to the dataset for iterative fine-tuning.}
    \label{fig:user_refinement}
\end{figure*}
\begin{figure*}[htb!]
    \centering
    \includegraphics[width=\linewidth]{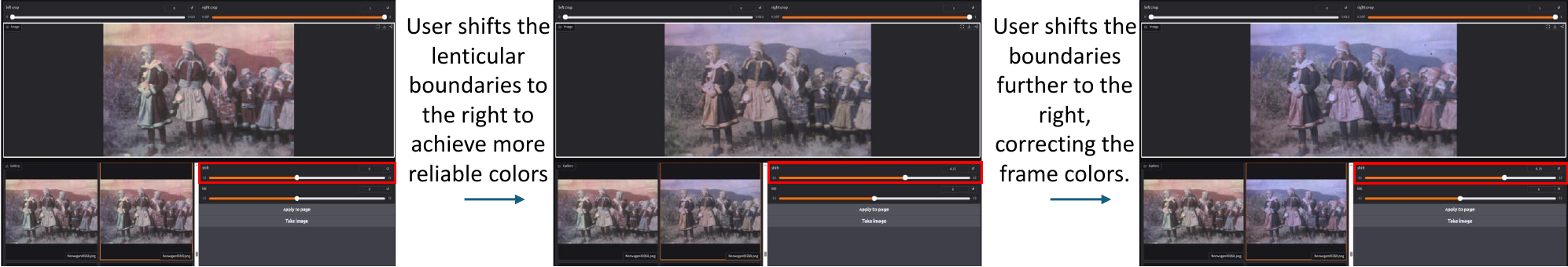}
    \caption{\textit{Shift lenticular boundaries}: (Left) The discolored film frame from deep-doLCE. (Middle) The user shifts the lenticular boundaries to the right to achieve more reliable colors. (Right) The user shifts the boundaries further to the right, correcting the frame colors.}
    \label{fig:dolce2_shift}
\end{figure*}
\begin{figure*}[htb!]
    \centering
    \includegraphics[width=\linewidth]{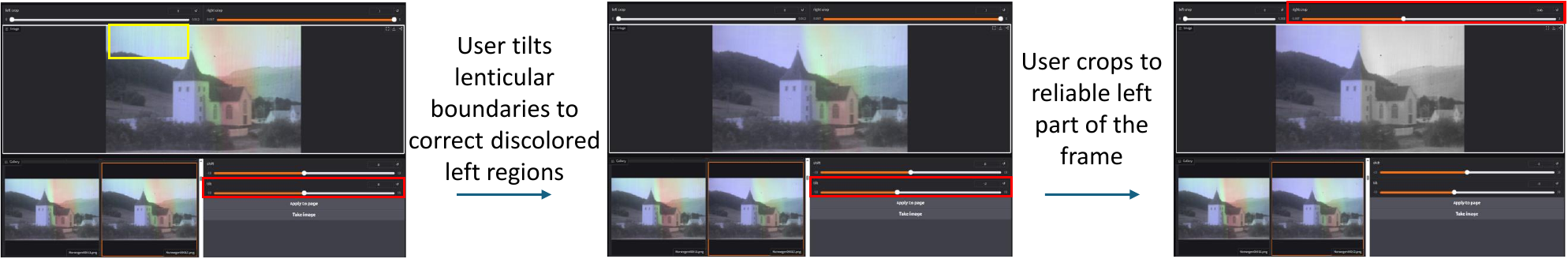}
    \caption{\textit{Tilt lenticular boundaries}: (Left) The discolored film frame from deep-doLCE. (Middle) The user tilts the lenticular boundaries to correct the discolored left region (highlighted with a yellow box), while the right region remains uncorrectable. (Right) The user then crops to the reliable corrected left part of the frame.}
    \label{fig:dolce2_tilt}
\end{figure*}
\subsubsection{Expert verification and feedback}
A domain expert, typically a film preservation specialist, then reviews and relabels the automated outputs frame by frame, interacting with each frame to decide which regions to accept, correct, or discard. If the colorization is already acceptable for the entire image, the expert simply accepts the frame as is. The user interface for expert-guided lenticular boundary annotation is presented in Figure~\ref{fig:hitl_ui}. All frames are organized into a browsable gallery and displayed in pages, so that only a small subset of frames is loaded at a time. This avoids the need to process or store the entire film sequence in memory at once, keeping the interface responsive even for long film sequences. Approximately 100 frames are loaded per page, allowing the expert to begin reviewing once the current page is loaded while the remaining frames are colorized in the background. Each frame is colorized on-the-fly using deep-doLCE~\cite{daronco2022deeplearningapproachdigital} as it is loaded into view, taking approximately 2--5 seconds per frame depending on GPU and scan resolution (e.g., roughly 3 seconds on an NVIDIA A100 40GB for the dataset presented in this paper). Once a frame is selected, the expert can apply three pixel-level operations to the lenticular boundaries: \textit{cropping}, \textit{tilting}, and \textit{shifting}. After adjusting the boundaries, the expert can accept the corrected frame for training and proceed to the next frame. The resulting annotations can be saved for subsequent fine-tuning.
\paragraph{Cropping/selecting reliable regions of interest}
When only part of a frame is reliably colorized, the expert can crop to the region of interest. As shown in Figure~\ref{fig:user_refinement}, the original frame contains discolored regions in the deep-doLCE output (left). The expert crops to the reliable part of the frame, with the excluded region visualized in grayscale to clearly indicate what will be discarded (middle). Finally, only the accepted region is added to the training dataset for iterative fine-tuning (right).
\paragraph{Shifting lenticular boundaries} When color errors arise from a global horizontal misalignment of the detected lenticular boundaries, the expert can shift the entire boundary grid left or right to realign it with the underlying lenticule pattern. As shown in Figure~\ref{fig:dolce2_shift}, the original frame exhibits discoloration due to an offset in the detected boundaries (left). The expert shifts the boundaries to the right, producing a visibly improved reconstruction (middle). A further shift fully corrects the frame colors (right).
\paragraph{Tilting lenticular boundaries} When color errors are caused by a global angular misalignment, the expert can rotate the lenticular boundary grid clockwise or counterclockwise to match the true orientation of the lenticules. As shown in Figure~\ref{fig:dolce2_tilt}, the original frame contains a discolored region on the left side (left). The expert tilts the boundaries counterclockwise to correct the affected region, highlighted by the yellow box (middle). Since the right portion of the frame remains uncorrectable, the expert then crops to the reliably reconstructed left region (right).
\paragraph{Discarding unrecoverable frames} Frames with unrecoverable issues, such as severe physical damage, extreme exposure problems, or fundamental detection failures, can be explicitly discarded. 
In summary, expert verification and feedback combines three complementary operations: cropping to reliable regions, shifting boundaries, and tilting boundaries. Together, these operations ensure that only trustworthy data is retained. Completely discolored frames are discarded, and partially correct frames contribute only their reliable regions. The result is a refined, high-quality dataset that enables more accurate and trustworthy AI-based color reconstruction. A video demonstration is available at \href{https://drive.google.com/file/d/1-oDvKvG6F-C938sQBCEnmMroOhZVsvo5/view?usp=sharing}{\textit{Expert verification and feedback Demo}}.
\subsubsection{Fine-tuning}
The corrected lenticule boundary positions constitute new, high-quality training labels for frames where the original automated detection was insufficient. These expert-corrected annotations are collected and used to fine-tune the lenticule detection network. The corrected frames and their updated boundary labels are added to the training set, and fine-tuning is always performed from the baseline model weights (rather than from previously fine-tuned checkpoints) with a reduced learning rate of $1 \times 10^{-5}$, in order to preserve the model's existing capabilities while adapting to the newly identified failure modes.
\subsubsection{Iterative re-processing and embedding of priors}
After each fine-tuning round, the updated model is re-applied to all frames, including those that previously failed or were discarded. Frames that were previously uncorrectable by the automated system may now be handled successfully by the improved model, as it has learned from similar, expert-corrected examples. The expert reviews the new outputs, and the cycle repeats until the desired quality level is achieved across the entire film. This iterative process typically converges within two to three cycles, as the corrections from early iterations address the most common failure modes and the fine-tuned model generalizes these corrections to similar frames throughout the film. Conceptually, this constitutes a recursive embedding of expert priors about valid lenticular boundaries that results in acceptable color appearance into the model. If all frames are deemed satisfactory by the expert, they are then post-processed as described in the following subsection.
\subsection{Post-processing}
\label{subsec:post_processing}
To better preserve the grain structure and fine detail of the original silver emulsion, we apply a simple post-processing step in CIELAB color space. The method is actually not specific to CIELAB. Any color space that separates brightness from chromatic information (e.g., YUV, HSV, or CIELAB) can be used to achieve the same effect. The key idea is to retain the lightness (and thus most of the texture and grain) from the scanned film, while using only the chromatic information from the learned colorization.

For each frame, we first convert both the original scanned film and the corresponding colorized output of our model to CIELAB space. For the scan (which is originally grayscale), we replicate the single channel into three channels before conversion and obtain its brightness component $L_{\text{scan}}$. For the model output, we obtain $L_{\text{pred}}, a_{\text{pred}}, b_{\text{pred}}$.

We then construct a fused Lab representation by taking the $L$ channel from the scan ($L_{scan}$) and the chromatic channels ($a_{pred}$, $b_{pred}$) from the model output. In practice, we optionally apply a frame-wise linear scaling to $a_{\text{pred}}$ and $b_{\text{pred}}$ so that their magnitude is matched to the scale of the input material, keeping the overall colorfulness plausible while avoiding over-saturation. This can be written as
\begin{equation}
    \tilde{a}_{\text{pred}} = s \, a_{\text{pred}}, \qquad
    \tilde{b}_{\text{pred}} = s \, b_{\text{pred}},
\end{equation}
where $s$ is a scalar chosen per film to account for the difference in resolution between the original scan and the network output (we set $s$ according to the ratio between the higher-resolution input scan and the lower-resolution model output).

The final Lab image is therefore given by
\begin{equation}
    (L, a, b) = \bigl(L_{\text{scan}}, \tilde{a}_{\text{pred}}, \tilde{b}_{\text{pred}}\bigr).
\end{equation}
This fused Lab image is then converted back to RGB to produce the final output frame. In this way, we preserve the texture and grain information from the original scan while benefiting from the color reconstruction provided by the model.

%% file: text/evaluation.tex
\section{Evaluation}
\label{sec:evaluation}
In this section, we evaluate the proposed human-in-the-loop framework on a historical lenticular film sequence and compare it against an automatic deep-learning baseline. We then analyze the impact of the Lab-based post-processing step described in Section~\ref{subsec:post_processing} on the final visual quality of the reconstructed frames.
\subsection{Dataset}
\label{subsec:dataset}
We conducted our evaluation on a historical lenticular film sequence titled \emph{Ein Besuch im Lappenlager} (``A visit to a Sami settlement''), filmed by a German amateur filmmaker in  1930 and preserved as part of an archive collection in Stuttgart, Germany.

 The lenticules are difficult to detect in several areas of the film scan; as a result, the fully automatic lenticular detection pipeline~\cite{daronco2022deeplearningapproachdigital} failed to reconstruct plausible colors for most frames, making this sequence a natural premiere use case for our human-in-the-loop method.
\begin{figure*}[htb!]
    \centering
    \includegraphics[width=\linewidth]{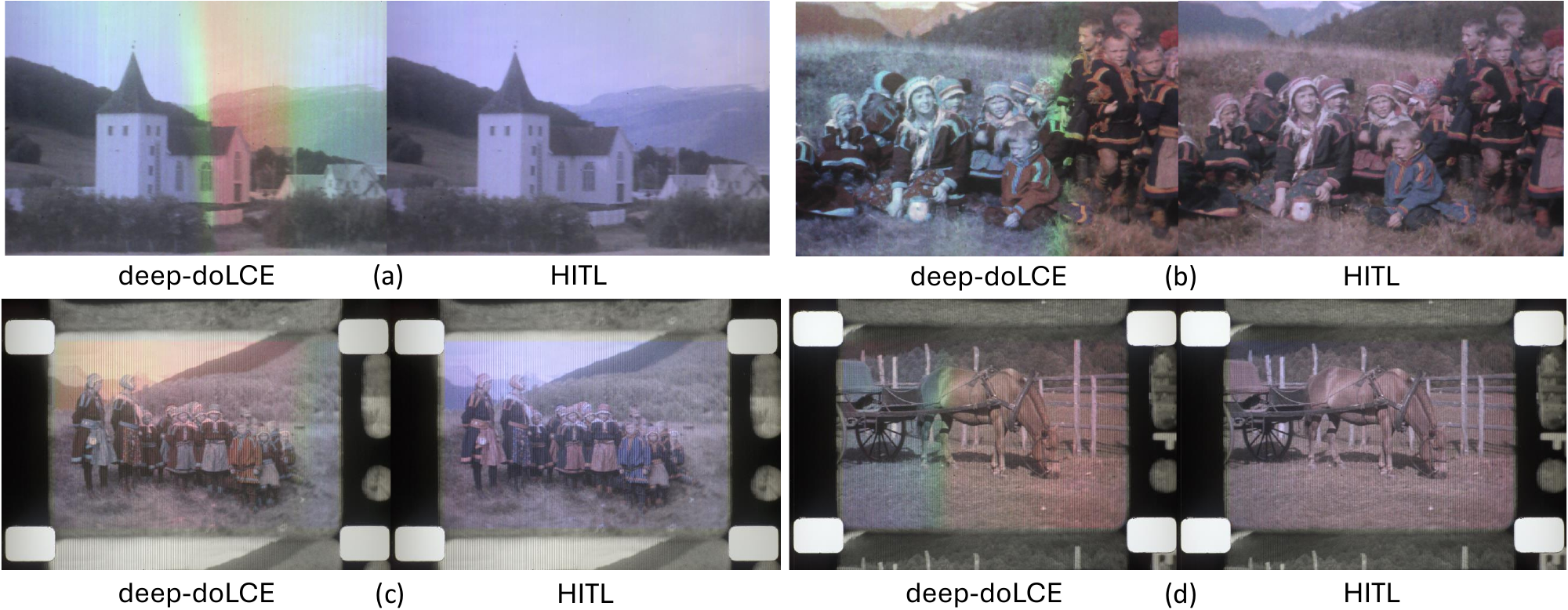}
   \caption{Qualitative comparison with the fully automatic color reconstruction pipeline (deep-doLCE)~\cite{daronco2022deeplearningapproachdigital} on the \emph{Ein Besuch im Lappenlager} sequence. Subfigures (a) and (b) show two representative frames, comparing the baseline deep-doLCE output with our human-in-the-loop reconstruction before Lab-space post-processing. Subfigures (c) and (d) show two additional frames after applying the lightness-preserving CIELAB post-processing step.}
    \label{fig:qualitative_dolce1}
\end{figure*}
\begin{figure*}[htb!]
    \centering
    \includegraphics[width=\linewidth]{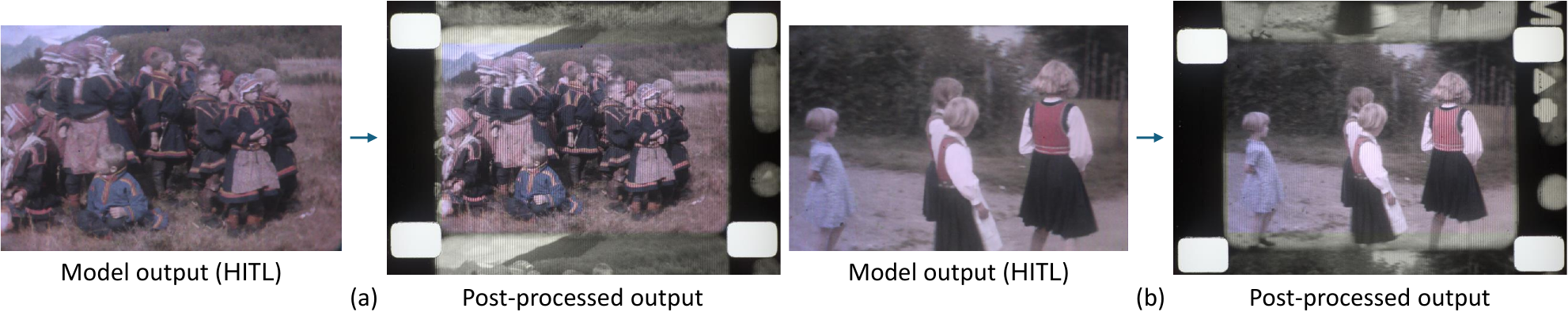}
   \caption{Effect of post-processing on reconstructed frames. The figure shows model outputs before and after post-processing for two representative frames ((a) and (b)). The post-processing step replaces the model's brightness with that of the original scan, preserving fine texture and grain while maintaining the reconstructed color. This results in improved visual fidelity and texture consistency in both examples.}
    \label{fig:qualitative_postprocessing}
\end{figure*}
\begin{figure*}[htb!]
    \centering
    \includegraphics[width=\linewidth]{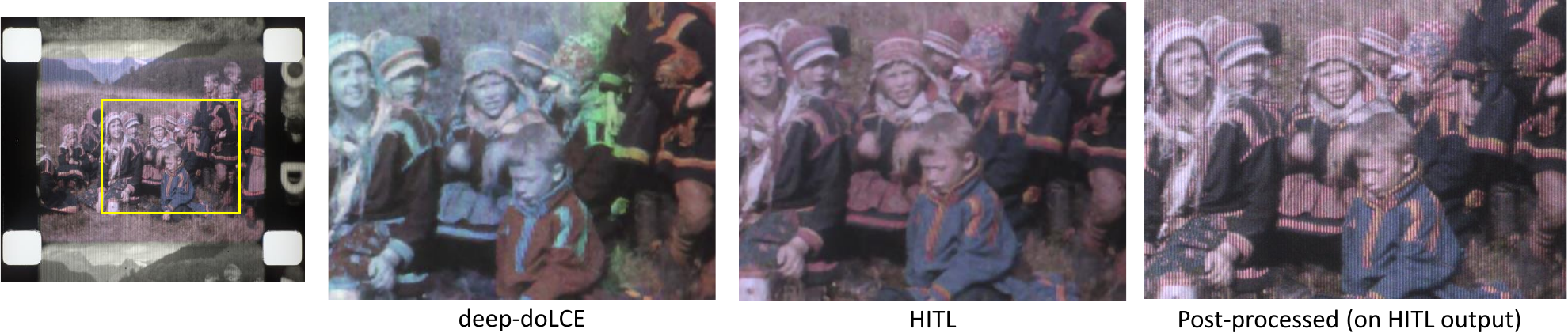}
   \caption{Zoomed-in comparison of deep-doLCE output (a), HITL output (b), and post-processed frame on HITL model output (c).}
    \label{fig:qualitative_zoomed_in_comparison}
\end{figure*}
\subsection{Comparison with Automated Color Reconstruction Pipeline}
We compare our human-in-the-loop framework with the fully automated
deep-doLCE color reconstruction pipeline~\cite{daronco2022deeplearningapproachdigital}
on the \emph{Ein Besuch im Lappenlager} sequence. Since no reliable color
ground truth is available, the comparison is based on qualitative visual
inspection. In the following subsections, we present a qualitative comparison of our method against the automated baseline, analyze the effect of iterative fine-tuning, and evaluate the impact of the post-processing step on the final reconstructed frames.
\subsubsection{Qualitative Comparison}
To show the potential of our approach, four representative failure cases
of the automated pipeline and the corresponding corrections achieved by our
method are presented in Figure~\ref{fig:qualitative_dolce1}. We selected
representative frames from different shots across the sequence so that the
examples cover a variety of discriminative visual features from the full
film. In all four examples, the visible artifacts in
deep-doLCE~\cite{daronco2022deeplearningapproachdigital} can be traced back
to local lenticular misalignment or incorrectly segmented boundaries.

In Figure~\ref{fig:qualitative_dolce1}(a), a church scene, the misalignment
leads to severe chroma instability with vertical rainbow-like bands cutting
through the sky and church. Our corrected boundaries remove these bands and
restore a spatially more uniform sky and landscape.

In Figure~\ref{fig:qualitative_dolce1}(b), with children in a field, the
automated result exhibits an abrupt left to right color split and a central
green streak due to a shift in the lenticular boundaries. By fine-tuning the
model with refined boundaries obtained by shifting the lenticular grid to
the right (see Figure~\ref{fig:dolce2_shift}), the model learns more
plausible boundary priors and produces consistent, natural colors across the
frame, eliminating the discontinuity entirely.

Figure~\ref{fig:qualitative_dolce1}(c) shows a group in traditional attire.
The automated pipeline produces a strong lateral color gradient from cold
gray-green to pink caused by mis-detected lenticules throughout the frame.
Our approach yields a balanced, uniform color palette with much reduced
discoloration. In Figure~\ref{fig:qualitative_dolce1}(d), with a horse and
cart, deep-doLCE introduces a prominent vertical green flare and blue
discoloration on the left due to misaligned lenticular boundary detection.
These artifacts disappear once the lenticular grid is corrected.

To further show the improvements achieved by our approach,
Figure~\ref{fig:qualitative_zoomed_in_comparison} provides a zoomed-in
comparison of the region highlighted by the yellow box in
Figure~\ref{fig:qualitative_dolce1}(c), showing a group of people including
a child wearing traditional S\'{a}mi attire.
Figure~\ref{fig:qualitative_zoomed_in_comparison}(a) shows the deep-doLCE
output, which appears largely decolorized, with the left half of the frame
particularly affected by a loss of chromatic information and a flat,
washed-out appearance. Figure~\ref{fig:qualitative_zoomed_in_comparison}(b)
shows the HITL reconstruction, which successfully restores color across the
frame, introducing rich bluish and magenta tones that recover the visual
character of the original scene. Background-to-foreground contrast is
enhanced and the color discontinuity visible in
Figure~\ref{fig:qualitative_zoomed_in_comparison}(a) is eliminated.
Figure~\ref{fig:qualitative_zoomed_in_comparison}(c) shows the
post-processed output, which builds on the HITL result by further enhancing
texture and fine surface details, producing the sharpest visual separation
and the most faithful rendering of the intricate patterns typical of
S\'{a}mi traditional wear. As the pipeline progresses from
Figure~\ref{fig:qualitative_zoomed_in_comparison}(a) to
Figure~\ref{fig:qualitative_zoomed_in_comparison}(c), the HITL refinement
and post-processing together recover the color, contrast, and texture that
the fully automated deep-doLCE reconstruction missed.
\begin{figure*}[htb!]
    \centering
    \includegraphics[width=\linewidth]{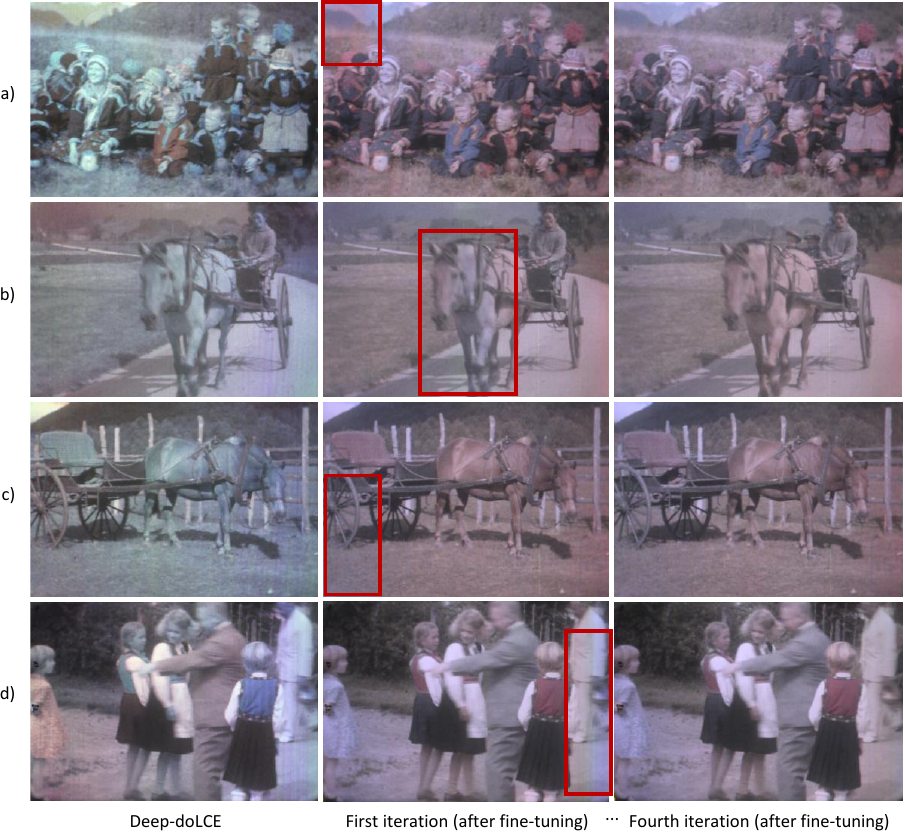}
   \caption{ Iterative fine-tuning progression on four representative frames (a--d). Each row shows the same frame across three stages: the baseline deep-doLCE output (left), the reconstruction after the first fine-tuning iteration with expert-corrected annotations (middle), and the reconstruction after the fourth fine-tuning iteration (right). After the first iteration, color reconstruction is substantially improved across all frames; however, small regions of residual discolorization remain (highlighted by red boxes). After the fourth iteration, these artifacts are mostly resolved and the color reconstruction appears plausible throughout all frames}
    \label{fig:fine-tuning}
\end{figure*}
\subsubsection{Effect of Iterative Fine-Tuning}
\label{subsubsec:effect_finetuning}
To show the effect of fine-tuning over different iterations of expert validated 
data, we present the progression of color reconstruction quality on the
\emph{Ein Besuch im Lappenlager} sequence. This sequence was particularly
challenging due to severe and widespread lenticular misalignment artifacts using both doLCE~\cite{ReutelerGschwind2014RundbriefFotografie} and deep-doLCE~\cite{daronco2022deeplearningapproachdigital},
making it a good case study for understanding the trends of iterative
refinement. We note that this is only one possible way to apply the
approach, and a broader evaluation across many different film sequences would be needed
to establish a more general best practice. Such an evaluation would require
a structured user study with domain experts, which we consider future work
and is beyond the scope of this paper.

To illustrate the effect of iterative fine-tuning,
Figure~\ref{fig:fine-tuning} presents the progression of color
reconstruction quality on four representative frames, where the left column
shows the deep-doLCE reconstruction, the middle column shows the result
after the first iteration, and the right column shows the final result after
the fourth iteration. Each round builds on the expert-corrected annotations
collected in the previous one. 

In the first iteration, approximately 265 frames (11\% of the sequence)
with plausible but imperfect reconstructions were selected and improved by
the expert using the full set of interactive operations, including shifting,
tilting, and cropping to reliable regions of interest. This round
substantially improved color reconstruction across all frames, as seen in
Figure~\ref{fig:fine-tuning} (middle column). However, small regions of
residual discolorization remained, as highlighted by the red boxes in
Figure~\ref{fig:fine-tuning} (middle column). In the second iteration,
expert corrections were mostly limited to cropping with only minor shifting in very few cases, as the model had already
learned to handle most misalignment cases from the first round and no
shifting or tilting was required. In the third iteration, only border
cropping was performed, reflecting the high reconstruction quality reached
after the first two rounds, with only marginal edge artifacts remaining. In
the fourth and final iteration, approximately 165 frames (7\% of the
sequence) were judged to be of very high quality and were accepted without
any corrective operation, being added directly to the training set. After
this iteration, residual artifacts were mostly resolved and the color
reconstruction appeared plausible throughout the entire sequence, as shown
in Figure~\ref{fig:fine-tuning} (right column).

For this sequence, the process shows a clear trend toward convergence, with
the most demanding corrections concentrated in the first iteration and
progressively less expert intervention required in subsequent rounds.
However, we note that this trend may depend on the specific sequence, the
condition of the film material, and the experience of the expert operator.
Further studies across a broader range of sequences are needed to draw more
general conclusions.
\subsubsection{Effect of post-processing}
The post-processing step preserves the fine texture, grain, and dynamic
range of the original film scan in the reconstructed frames. As illustrated
in Figure~\ref{fig:qualitative_postprocessing}, regions with fine details
such as foliage, fabric patterns, and facial features benefit noticeably,
with improved clarity and texture consistency. This approach combines the
strengths of learned colorization with the intrinsic qualities of the
scanned film, resulting in outputs that are visually sharper. This is confirmed in Figure~\ref{fig:qualitative_zoomed_in_comparison}, which provides a
zoomed-in comparison of the deep-doLCE output
(Figure~\ref{fig:qualitative_zoomed_in_comparison}(a)), the HITL output
(Figure~\ref{fig:qualitative_zoomed_in_comparison}(b)), and the
post-processed result (Figure~\ref{fig:qualitative_zoomed_in_comparison}(c)).
The zoomed view shows that while the HITL reconstruction successfully
restores color, the post-processed output additionally recovers fine surface
details, film grain, and local texture that are characteristic of the
original scan.

%% file: text/conclusion.tex
\section{Conclusion}
\label{sec:conclusion}
We presented a human-in-the-loop (HITL) deep learning framework for color reconstruction of historical lenticular films. By decoupling lenticular boundary detection from colorization and exposing an editable intermediate representation, our pipeline enables domain experts to interactively correct failure cases through shifting, tilting, and cropping operations that fully automated methods such as deep-doLCE cannot reliably handle. Expert-corrected annotations are used to iteratively fine-tune the detection model, progressively embedding domain knowledge and improving robustness across the entire film sequence. On the challenging \emph{Ein Besuch im Lappenlager} sequence, this iterative process converged within four rounds, with the most intensive corrections concentrated in the first iteration and decreasing expert effort required in each subsequent round. A final post-processing step replaces the model brightness with that of the original scan, ensuring that reconstructed frames retain the authentic texture, grain, and dynamic range of the historical material, which is a critical requirement for archival and presentation purposes. Together, the HITL refinement and post-processing stages produce results that substantially outperform the fully automated baseline in both color fidelity and visual authenticity, as confirmed by qualitative analysis across representative frames and zoomed-in comparisons.

This work opens several promising directions for future research. A structured user study with domain experts across a broader range of film sequences is needed to establish best practices, in particular to quantify the relationship between the number of annotated frames and reconstruction quality, providing practical guidance on annotation effort. Furthermore, a dedicated performance study of the tool is needed to understand its requirements across different hardware configurations. Currently, approximately 100 frames are loaded per page, which is reasonable for high-powered GPUs but may require adaptation for less capable systems. Adopting a distributed architecture for frame loading and colorization could significantly reduce wait times and increase the practicality of the tool for large-scale archival workflows. Additionally, federated or continual learning strategies could allow model updates fine-tuned by different experts on different sequences to be aggregated, moving toward a more universal lenticular reconstruction model applicable across diverse film stocks and capture conditions. Finally, extending the evaluation to a wider variety of lenticular processes beyond Kodacolor would broaden the applicability of the framework to the full range of historically significant color film formats with spatial encoding on silver emulsion.

%% file: text/acknowlegements.tex
\subsection*{Author Contributions}
\textbf{Saptarshi Neil Sinha:} Conceptualization, Methodology (Human-in-the-loop pipeline), Software, Validation, Formal Analysis, Investigation, Data Curation, Visualization, Writing - Original Draft, Writing - Review \& Editing, Supervision, Project administration, Validation, Resources. \textbf{Tiago Kleist:} Methodology (Human-in-the-loop and post-processing pipeline), Software, Validation, Investigation, Data Curation, Visualization, Writing - Original Draft, Writing - Review \& Editing,  \textbf{Giorgio Trumpy:} Methodology (Post-processing pipeline), Validation, Writing - Original Draft, Writing - Review \& Editing, Supervision, Project administration.

\subsection*{Acknowledgments}
The work presented in this paper has been partially funded by the European Commission through the Horizon Europe projects PERCEIVE under Grant Agreement No. 101061157 and COLOURS under Grant Agreement No. 101233413. The authors wish to thank Anna Leippe (Haus des Dokumentarfilms, Stuttgart) for providing the case study and conducting the historical research. They are also grateful to Joachim Reuteler for his technical support. Rudolf Gschwind (Emeritus, University of Basel) contributed to the early stages of this research, and this work builds upon many of his pioneering studies in the field of digital imaging for color photography.

\subsection*{Financial Disclosure}

None reported.

\subsection*{Conflicts of Interest}

The authors declare no conflicts of interest.